\documentclass{article}

\PassOptionsToPackage{numbers, compress}{natbib}

\usepackage[preprint]{neurips_2024}

\usepackage[utf8]{inputenc} 
\usepackage[T1]{fontenc}    
\usepackage{hyperref}       
\usepackage{url}            
\usepackage{booktabs}       
\usepackage{amsfonts}       
\usepackage{nicefrac}       
\usepackage{microtype}      
\usepackage{xcolor}         

\usepackage{amsmath} 
\usepackage{amssymb}  
\usepackage{graphicx}
\usepackage{multirow}
\usepackage{bbding}
\usepackage{enumitem}
\usepackage{tcolorbox}
\usepackage{subfig}
\usepackage{stfloats}
\usepackage{natbib}
\usepackage{tabularx}

\usepackage{amsmath,amsfonts,bm}

\def\eqref#1{equation~\ref{#1}}

\def\1{\bm{1}}

\def\va{{\bm{a}}}

\def\vm{{\bm{m}}}

\def\vq{{\bm{q}}}

\def\vI{{\bm{I}}}
\def\vP{{\bm{P}}}
\def\vS{{\bm{S}}}
\def\vpi{{\bm{\pi}}}
\def\vPi{{\bm{\Pi}}}

\DeclareMathAlphabet{\mathsfit}{\encodingdefault}{\sfdefault}{m}{sl}
\SetMathAlphabet{\mathsfit}{bold}{\encodingdefault}{\sfdefault}{bx}{n}

\newcommand\term{composable generalization}

\newcommand\TeRm{Composable Generalization}
\newcommand\tabbre{CGA}
\newcommand\data{RH20T-P}
\newcommand\agent{RA-P}
\newcommand{\VarSty}[1]{\textnormal{\ttfamily\color{blue!90!black}#1}\unskip}

\title{\data{}: A Primitive-Level Robotic Manipulation Dataset towards \TeRm{} Agents \\in Real-world Scenarios}

\makeatletter
\newcommand{\printfnsymbol}[1]{%
  \textsuperscript{\@fnsymbol{#1}}%
}
\makeatother

\author{
    Zeren Chen\textsuperscript{\rm 1,2}\thanks{Equal contribution.},\;
    Zhelun Shi\textsuperscript{\rm 2}\printfnsymbol{1},\;
    Xiaoya Lu\textsuperscript{\rm 1,3}\printfnsymbol{1},\;
    Lehan He\textsuperscript{\rm 2}\printfnsymbol{1},\;
    Sucheng Qian\textsuperscript{\rm 3}, \\
\textbf{
    Zhenfei Yin\textsuperscript{\rm 1,4}, \;
    Wanli Ouyang\textsuperscript{\rm 1,4}, \;
    Jing Shao\textsuperscript{\rm 1}\thanks{Corresponding author.}, \;
    Yu Qiao\textsuperscript{\rm 1}, \;
    Cewu Lu\textsuperscript{\rm 3}\printfnsymbol{2}, \;
    Lu Sheng\textsuperscript{\rm 2}\printfnsymbol{2}
} \\ \\
    \large\textsuperscript{\rm 1} Shanghai AI Laboratory, \;
    \large\textsuperscript{\rm 2} School of Software, Beihang University,\\
    \large\textsuperscript{\rm 3} Shanghai Jiao Tong University, \;
    \large\textsuperscript{\rm 4} University of Sydney \\ \\
    \small\texttt{\{czr1604,shizhelun,lsheng\}@buaa.edu.cn},\; \small\texttt{shaojing@pjlab.org.cn}
}

\begin{document}

\makeatletter
\let\@oldmaketitle\@maketitle%
\renewcommand{\@maketitle}{\@oldmaketitle%
    \centering
    \vspace*{1mm}
    \includegraphics[width=0.99\linewidth]{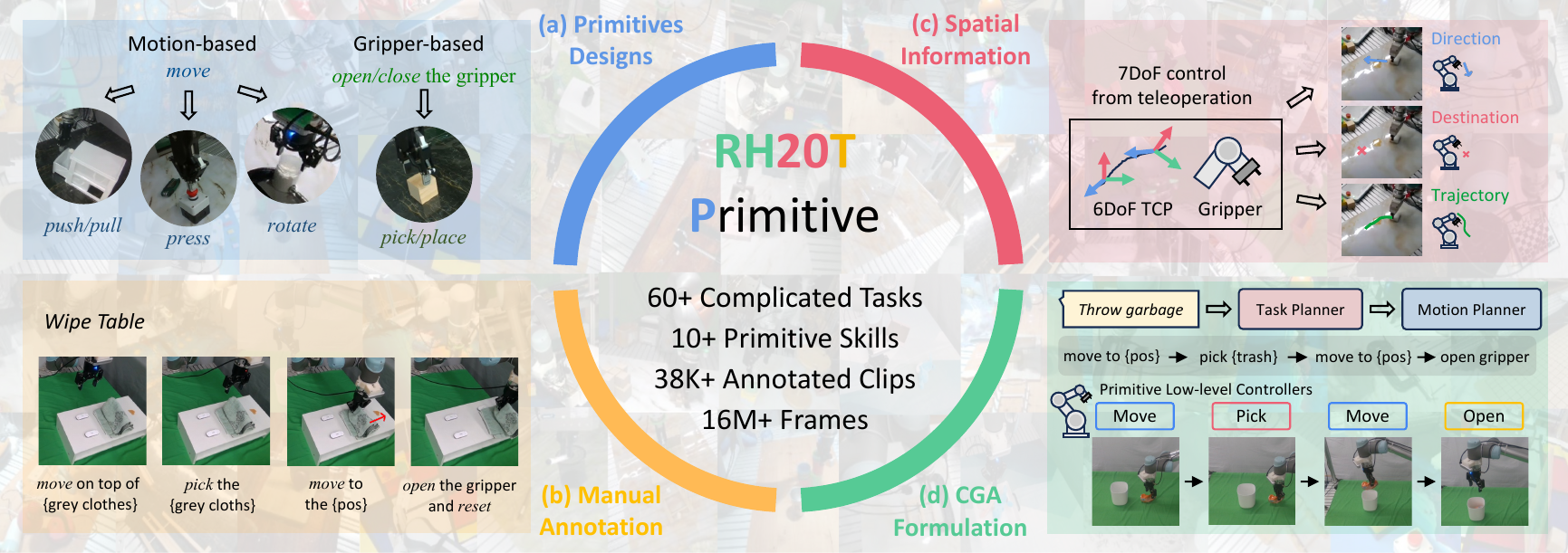}
    \captionof{figure}{
        \textbf{Overview of our \data{} dataset.}
        We propose a primitive-level robotic manipulation dataset for future development of \term{} agents. 
        With meticulously-defined primitive skills and multiple forms of spatial information, each manipulation episode is carefully annotated manually.
        We also standardize a plan-execute paradigm for \tabbre{}s based on \data{}, which can generalize to novel skills through \term{}.
    }
    \label{fig:overview}
}
\makeatother
\maketitle

\begin{abstract}
Achieving generalizability in solving out-of-distribution tasks is one of the ultimate goals of learning robotic manipulation.
Recent progress of Vision-Language Models (VLMs) has shown that VLM-based task planners can alleviate the difficulty of solving novel tasks, by decomposing the compounded tasks as a plan of sequentially executing primitive-level skills that have been already mastered.
It is also promising for robotic manipulation to adapt such composable generalization ability, in the form of \term{} agents (\tabbre{}s).
However, the community lacks of reliable design of primitive skills and a sufficient amount of primitive-level data annotations.
Therefore, we propose \data{}, a primitive-level robotic manipulation dataset, which contains about 38k video clips covering 67 diverse manipulation tasks in real-world scenarios. 
Each clip is manually annotated according to a set of meticulously designed primitive skills that are common in robotic manipulation.
Furthermore, we standardize a plan-execute \tabbre{} paradigm and implement an exemplar baseline called \agent{} on our \data{}, whose positive performance on solving unseen tasks validates that the proposed dataset can offer composable generalization ability to robotic manipulation agents. 
Project homepage: \url{https://sites.google.com/view/rh20t-primitive/main}.
\end{abstract}

\section{INTRODUCTION}
\label{sec:intro}

Robotic manipulation tasks are designed to enable robotic systems to comprehend environmental observations and open-world task instructions like language, guiding actuators to execute specific actions in real-world applications.
Previous attempts at imitation learning~\cite{brohan2022rt1,shridhar2022cliport,zhao2023act,jang2022bc} or reinforcement learning~\cite{escontrela2024video,luo2023action,hansen2022modem,adeniji2023language} often struggle to generalize to out-of-distribution tasks.
Since collecting all real-world tasks to build in-distribution training datasets is impractical, crafting generalizable robotic manipulation agents in this fashion poses challenges.

Recently, Vision-Language Models (VLMs)~\cite{openai2023gpt4v,liu2024llava} have shown impressive potential in following multimodal instructions.
Some approaches~\cite{brohan2023rt2} fine-tune VLMs by aligning the VLMs' language decoders (such as PaLM-E~\cite{driess2023palme}) with the distribution of action sequences in robotic manipulation tasks.
While to some extent, they can generalize to tasks involving novel objects or novel compositions between known skills and known objects, the skill set is restricted to those encountered in the training datasets.
Alternatively, other methods~\cite{driess2023palme,hu2023vila,wake2023gpt4robotics} employ VLMs as task planners, breaking down a compounded task-solving procedure into a sequence of primitive-level skill-execution subroutines.
We formulate this promising paradigm as \textit{\textbf{\term{}}}, which facilitates that, as shown in Figure~\ref{fig:overview} (d), a novel skill ``throw'' that never presents in the training data can be accomplished by executing more straightforward and common skills, such as \{\textit{move}, \textit{pick}, \textit{move}, \textit{open}\} in composition. 
We argue that the \term{} agents (\tabbre{}s) that follow this paradigm can mitigate the unpredictability and intricacy when encountering out-of-distribution compounded robotic manipulation tasks.


Existing \tabbre{}s~\cite{hu2023vila,wake2023gpt4robotics} primarily focus on planning systems with off-the-shelf VLMs like GPT-4V~\cite{openai2023gpt4v}, while the research on entire \tabbre{} system remains insufficient.
Particularly, \textbf{\textit{how to predict reliable primitive-level spatial information that grounds successful executions of primitive skills?}}
We refer it as primitive-level motion planning.
For example, identifying virtual points in 3D space for generating robust trajectories without touching obstacles.
VLMs struggle to provide such necessary primitive-level spatial information.
While these agents can delegate motion planning to low-level controllers or integrate an additional motion planner, a lack of primitive-level spatial knowledge in current robotic manipulation datasets makes it hard for acquiring specialized controllers or motion planners, resulting in low execution success rates in more compounded tasks.
Thus, we are motivated to collect \textbf{\data{}}, a robotic manipulation dataset built on \textbf{RH20T}~\cite{fang2023rh20t} at a \textbf{P}rimitive level, with meticulously designed primitive skills and diverse primitive-level spatial knowledge, making it feasible to construct generalizable \tabbre{}s in real-world scenarios.

In \data{}, we design a set of hierarchical and scalable primitive skills based on two types of skills, \emph{i.e.}, motion-based and gripper-based skills.
Each motion-based skill is equipped with various forms of spatial knowledge like trajectories.
Next, manipulation episodes in \data{} are manually segmented accordingly (about 38k clips covering 67 tasks).
Additionally, we standardize a plan-execute \tabbre{} paradigm, as shown in Figure~\ref{fig:overview} (d), 
and implement an exemplar baseline \tabbre{} on \data{}, called \textbf{\agent{}} (\textbf{R}obot\textbf{A}gent-\textbf{P}rimitive).
Our \agent{} showcases feasibility and generzalization in real-world demonstrations, even on novel skills, validating the composable generalization ability offered by \data{}.
We believe that the \data{} dataset will pave the way for the development of more potent \tabbre{}s in the future.

\section{RELATED WORK}
\label{sec:related_work}

\begin{table*}[tb]
  \caption{\textbf{Comparison with Existing Robotic Manipulation Dataset.}}
  \label{tab:data_comparison}
  \centering
  \resizebox{0.88\linewidth}{!}{
  \begin{tabular}{@{}lcccccc@{}}
    \toprule
    \multirow{2}{*}{Dataset} & \multirow{2}{*}{Amount} & \multicolumn{2}{c}{Modality} & \multicolumn{3}{c}{Annotation}  \\
    \cmidrule(lr){3-4}\cmidrule{5-7}
    & & Image & Depth & Action Seq. & Task Desc. & Plan Segmentation \\
    \midrule
    Bridge \cite{ebert2021bridge} & 7.2k & multi-view & \XSolidBrush & \Checkmark & \Checkmark & \XSolidBrush \\
    Open-x-embodiment \cite{padalkar2023open} & 1M & multi-view & \Checkmark & \Checkmark & \Checkmark & \XSolidBrush \\
    Droid \cite{khazatsky2024droid} & 76k & multi-view & \Checkmark & \Checkmark & \Checkmark & \XSolidBrush \\
    Lang-cond \cite{lynch2020language} & 50k & multi-view & \XSolidBrush & \Checkmark & \Checkmark & free-form language \\
    Language Table \cite{lynch2023languagetable} & 594k & single-view & \XSolidBrush & \Checkmark & \Checkmark & free-form language \\
    RH20T-P (ours) & 38k & multi-view & \Checkmark & \Checkmark & \Checkmark & pre-defined primitives \\
    \bottomrule
  \end{tabular}}
\end{table*}

\noindent\textbf{Vision-Language Models (VLMs).}
Vision-Language Models (VLMs) have gained significant attention due to their multimodal perception capabilities. 
Some studies~\cite{liu2024llava,dai2023instructblip,chen2023shikra,peng2023kosmos,lamm2023yin} incorporate image semantics into language models, dedicated to understanding 2D images. 
Among these, LLaVA~\cite{liu2024llava} adopts a two-stage instruction-tuning pipeline for general-purpose visual-language understanding. 
There are also some studies~\cite{xu2023pointllm,chen2023octavius,hong20243dllm} on VLMs in 3D vision.
For example, 3D-LLM~\cite{hong20243dllm} introduces the 3D semantics into LLMs by rendering point clouds into 2D images, enabling the models to perform a range of 3D tasks.

\vspace{+0.5mm}
\noindent\textbf{VLMs as Task Planners.}
Applying VLMs for planning~\cite{driess2023palme, hu2023vila, wake2023gpt4robotics, qin2023mp5} in robotic tasks have shown great potential.
PaLM-E~\cite{driess2023palme} develops an embodied VLM and trains it jointly on web-scale datasets, but the unavailability of primitive skills makes the granularity of output actions changing.
VILA~\cite{hu2023vila} and GPT4Robotics~\cite{wake2023gpt4robotics} conduct ICL with GPT-4V~\cite{openai2023gpt4v}, to generate primitive skills.
The proprietary nature of GPT-4V leads to extensive prompt engineering and inflexibility to utilize other multi-modal observations, \emph{e.g.}, depth. 

\vspace{+0.5mm}
\noindent\textbf{Primitive-level Robotic Manipulation Datasets.}
Using VLMs as planners for task decomposition has urgently increased the demand for primitive-level datasets tailored for \tabbre{}s.
We provide a comparison with existing robotic manipulation dataset in Table~\ref{tab:data_comparison}.
Most robotic manipulation datasets~\cite{fang2023rh20t, mandlekar2018roboturk, dasari2019robonet, sharma2018multiple, jang2022bc, ebert2021bridge, padalkar2023open} either lack textual annotations or only provide language descriptions for entire tasks, which are insufficient for \tabbre{}s.
While several datasets~\cite{lynch2023languagetable,nair2022learning,lynch2020language} feature hindsight free-form language for manipulation video clips, these coarsely-grained decompositions may confuse the low-level controllers.
The absence of robotic manipulation datasets with fine-grained primitive skills hinders the development of \tabbre{}s.

\section{\data{}: A Primitive-level Robotic Manipulation Dataset}
\label{sec:rh20tp}

\begin{figure*}[tb]
    \centering
    \includegraphics[width=\linewidth]{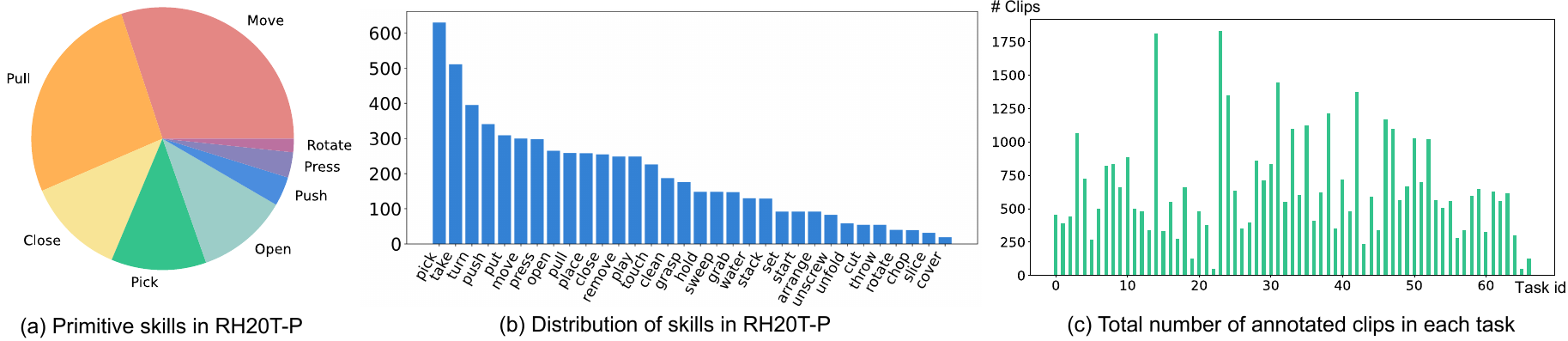}
    \caption{
        \textbf{Data statistics of \data{} dataset.}
    }
    \label{fig:rh20tp_stats}
\end{figure*}

\subsection{Preliminary of RH20T and Data Sampling}
\label{sec:preliminary_rh20t}

RH20T\footnote{https://rh20t.github.io, MIT license.}{~\cite{fang2023rh20t}}, data source of \data{}, encompasses a broad spectrum of real-world robot manipulation demonstrations, with each episode featuring diverse contexts, camera viewpoints, and language descriptions.
Its diversity and executability are pivot components for the development of intelligent robotic agents.
This motivates us to conduct primitive-level annotations for RH20T.
Specifically, we sample a subset of tasks in RH20T that are suitable for \tabbre{}s (\emph{e.g.}, visual reasoning) to construct a primitive-level robotic manipulation dataset.
The dataset still retains numerous complex skills beyond pick-and-place skills after sampling, such as wiping the table with a sponge or arranging pieces on a chessboard to complete the initial setup.
%

\subsection{Hierarchical and Scalable Primitive Skills}
\label{subsec:primitive_skills}

Primitive skills are crucial for \data{} dataset, as the way we decompose tasks shapes the design of the \tabbre{}s paradigm.
Applying free-form natural language~\cite{lynch2023languagetable,nair2022learning,lynch2020language} as primitive skills are overly coarse-grained, increasing difficulty for controllers to grasp task semantics, deviating from the original intent of \tabbre{}s.
Formally, we define primitive skills from the perspective of the robot arms, focusing on the state changes that primarily occur in the robot arm's motion and gripper during the manipulation process.
As shown in Figure~\ref{fig:overview} (a), we can divide the primitive skills into two categories: \textit{\textbf{motion-based}} and \textit{\textbf{gripper-based}} skills.

\begin{figure*}[tb]
    \centering
    \includegraphics[width=\linewidth]{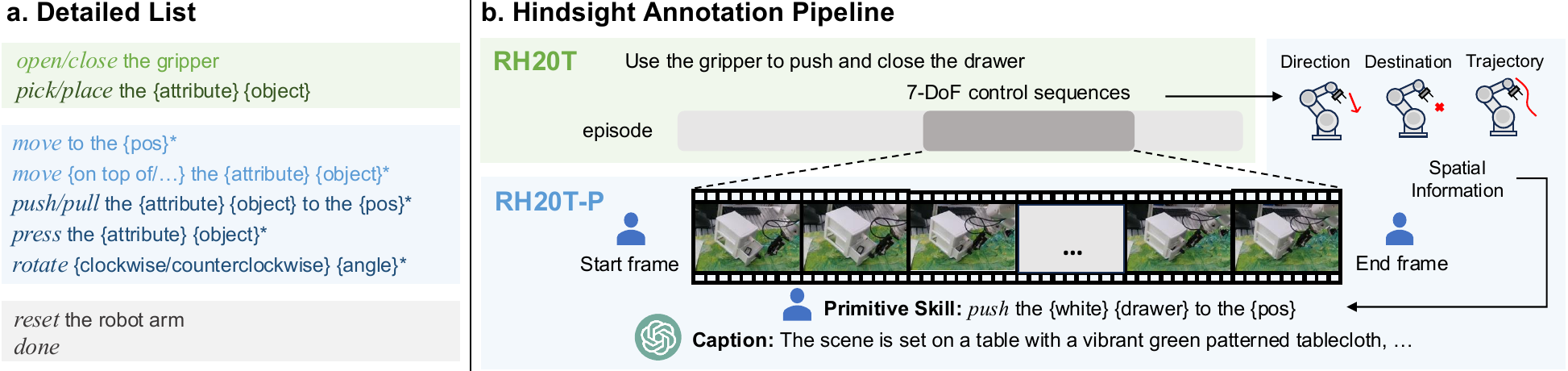}
    \caption{
        \textbf{(a) Detailed list of primitive skills in \data{}.} 
        ``*'' indicates this primitive skill contains spatial information of multiple form, \emph{e.g.}, destination or trajectory.
        \textbf{(b) Process of hindsight primitive-level annotation.} 
    }
    \label{fig:skills_and_anno}
\end{figure*}

\vspace{+0.5mm}
\noindent\textbf{Motion-based Skills.}
Motion-based skills are designed to describe each movement of robot arms.
To clearly characterize their behaviors beyond language, we equip each motion-based skill with corresponding spatial information extracted from teleoperation records in RH20T dataset.
These spatial information can be represented in multiple forms, such as reaching a destination or following a trajectory.
In contrast, primitive skills defined in VILA~\cite{hu2023vila} and SayCan~\cite{ahn2022saycan} lack precise spatial information, which may confuse low-level controllers when executing ambiguous primitive skills like ``move forward''.
Moreover, this deficiency often leads to inconsistency in granularity when executing subsequent primitive skills, \emph{e.g.}, a succeeding ``pick'' might necessitate an extra long movement to compensate for previous imprecise positioning, diverting the focus of low-level controllers from interacting with diverse objects.   

In our design, we define a set of hierarchical motion-based skills.
Among them, \textit{move} is the most foundational and versatile motion-based skill, encapsulating all types of motion.
Building on \textit{move}, we further define more specialized motion-based skills, such as \textit{pull} and \textit{press}, with complex and context-specific semantics in various scenarios.
It allows for tailored motion planning for different motion-based skills, \emph{e.g.}, pushing the object following a certain direction or moving the robot arm along a specific trajectory.
Also, more proficient controllers can be assigned to execute them.
Moreover, with such hierarchical definition, we can also easily expand more specialized semantics based on basic \textit{move} to tackle more challenging and novel tasks in the future.

\vspace{+0.5mm}
\noindent\textbf{Gripper-based Skills.}
Gripper-based skills are intuitive.
They include operations related to the gripper, such as \textit{pick}, which requires first opening the gripper and then closing.

\vspace{+0.5mm}
\noindent\textbf{Primitive Skill Template.}
To interact with different objects and attributes more flexibly, we design a primitive skill template, \emph{e.g.}, ``\textit{pick} the \{attribute\} \{object\}''. 
The placeholders in templates are completed during annotation, serving as the ground truth in \data{}.
The task planners in \tabbre{}s can perceive generalizable object categories and attributes in templates with their broad knowledge.
Such design can avoid situations like SayCan~\cite{ahn2022saycan} where defining all skill-object composition as primitive skills exhaustively.

\vspace{+0.5mm}
\noindent\textbf{All Primitive Skills.} 
Besides motion-based and gripper-based skills, we also define \textit{done}/\textit{reset} to indicate the task has been completed and the robot arm is required to be reset.
All primitive skills in \data{} are listed in Figure~\ref{fig:skills_and_anno} (a).

\subsection{Hindsight Primitive-level Annotation}
\label{subsec:anno}

We first ask the annotators to watch complete episode and segment each episode into video clips.
As shown in Figure~\ref{fig:skills_and_anno} (b), each of video clip is annotated with start frame and end frame, as well as corresponding primitive skills.
The placeholders for objects and attributes in the templates are annotated based on the video.
We then use the teleoperation records ($7$-DoF parameters) in RH20T dataset to generate the various forms of primitive-level spatial information (\emph{i.e.}, destination, direction, trajectory) for motion-based skills.
Besides, we use GPT-4V~\cite{openai2023gpt4v} to caption each episode, providing detailed descriptions for each scene.
Most of the hallucinations in these captions are removed after human inspection.
We also include these captions in \data{}.

\section{Plan-execute CGA Paradigm}
\label{sec:agent}

We standardize a plan-execute \tabbre{} paradigm, with two planners for task decomposition and motion planning, as well as primitive-level controllers for subsequent execution.

\begin{figure}[tb]
    \centering
    \includegraphics[width=\linewidth]{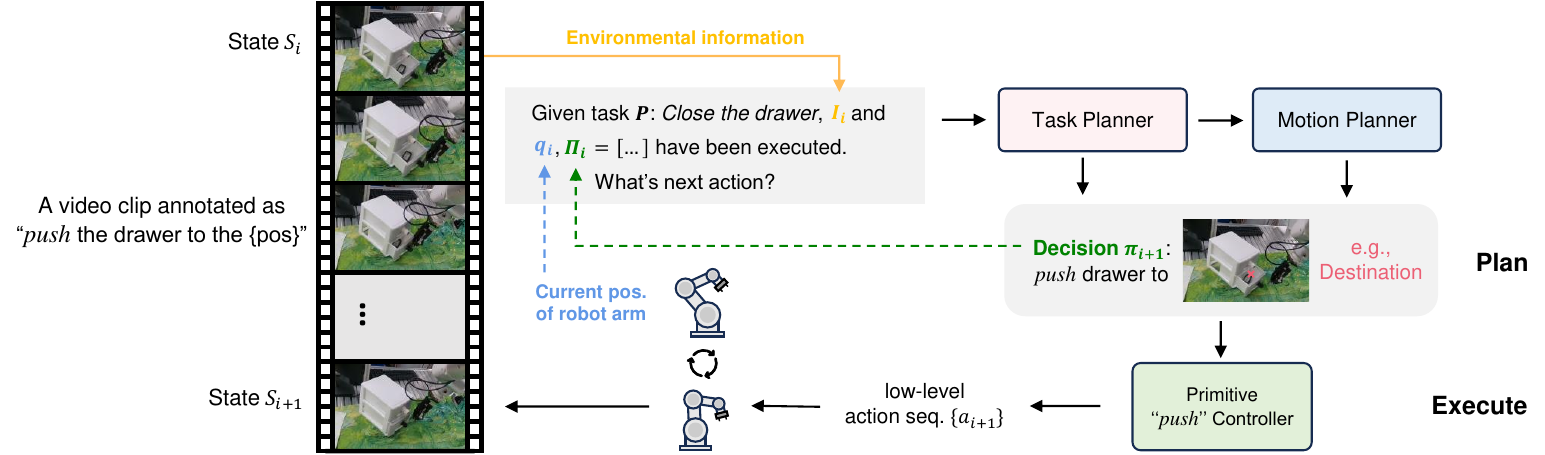}
    \caption{
        \textbf{Plan-execute CGA paradigm.}
    }
    \label{fig:agent_paradigm}
\end{figure}

\subsection{Overall Pipeline}
\label{subsec:pipeline}

As shown in Figure~\ref{fig:agent_paradigm}, given a language description $\vP$, we first define a state $\vS_i$, where the agent has completed a part of task from initial state $\vS_0$.
Based on historical decisions $\vPi_i=\{\vpi_0,\cdots,\vpi_i\}$ and observations under $\vS_i$, the task planner in \agent{} predict the next primitive-level decision $\vpi_{i+1}=\text{TaskPlanner}(\vP,\vPi_i,\vI_i,\vq_i)$.
Here, we collect visual features $\vI_i$ from cameras and the position of the robot arm $\vq_i$ as the input observation.
If output decision $\vpi_{i+1}$ belongs to motion-based skills, we further utilize motion planner to predict the corresponding primitive-level spatial information $\vm_{i+1}=\text{MotionPlanner}(\vpi_{i+1},\vI_i)$.
Next, the low-level controller maps the decision $\vpi_{i+1}$ and spatial information $\vm_{i+1}$ to action sequences $\{\va_{i+1}\} = \text{Controller}(\vpi_i,\vm_{i+1},\vI_i)$.
These action sequences are performed by the robot arm to transition from $\vS_i$ to the next state $\vS_{i+1}$.
Then a new round of planning is conducted under $\vS_{i+1}$.
In this manner, planner and controller continue to work alternately until planner gives a \textit{done} decision, indicating the task has been completed.

In practice, we associate each transition $\vS_i\xrightarrow{}\vS_{i+1}$ with a corresponding video clip from \data{} during training, based on the assumption that the part of task before $\vS_i$ has been successfully executed.
In each clip, we collect RGB images and position of the robot arm from the initial frame as the input observations $(\vI_i,\vq_i)$, expecting the task planner to make decisions $\vpi_{i+1}$ that are consistent with the primitive skill annotated in the clip.
The $7$-DoF control information recorded in the clips can be regarded as the action sequences $\{\va_{i+1}\}$ predicted by the low-level controller.
During the inference stage, we follow the above paradigm to sequentially conduct planning and execution step by step.

\subsection{RA-P: A Baseline Implementation}
\label{subsec:impl_details}

We implement a baseline \tabbre{}, \emph{i.e.}, \textbf{\agent{}}, on \data{}.

\vspace{+0.5mm}
\noindent\textbf{Task Planner.} 
We employ LLaVA\footnote{https://github.com/haotian-liu/LLaVA, Apache-2.0 license.}{~\cite{liu2024llava}} as task planner in \agent{}.
To fine-tune the language model, we also generate an instruction-following dataset with robotic manipulation knowledge based on \data{}.
Note that using other VLMs as task planner is feasible, \emph{e.g.}, applying GPT-4V via ICL.

\vspace{+0.5mm}
\noindent\textbf{Motion Planner.}
Various forms of spatial information in \data{} offer a broader range of options for motion planner.
We use destination $(x,y,d)$ of the trajectory as spatial information $\vm_i$ for simplicity, where $x,y$ denote the pixel coordinates in the image, and $d$ denotes the depth relative to the camera.
Here, we employ a Deformable DETR\footnote{https://github.com/fundamentalvision/Deformable-DETR, Apache-2.0 license.}{~\cite{zhu2020deformabledetr}} as a simple motion planner to localize next destination $(x,y,d)$ that the robot arm should move to.
Inspired by \cite{dai2021updetr,chen2023siamese,zang2023contextdet}, we introduce a special token \texttt{<pos>} to VLM vocabulary.
Once the \texttt{<pos>} is included in the prediction of task planner (\emph{e.g.}, "\textit{move} on top of the block \texttt{<pos>}"), the motion planner will be activated.
We then add the hidden features of the token \texttt{<pos>} with semantics related to objects and its spatial information to the object queries in DETR so that the DETR can localize the relevant destination.
Finally, we convert $(x,y,d)$ into a 3D point in real world with camera calibration for subsequent execution.

\vspace{+0.5mm}
\noindent\textbf{Low-level Controller.}
We apply two types of controllers, \emph{i.e.}, hard-code and policy-based controller.
Primitive skills like \textit{move} and \textit{open} can be executed directly or based on motion planning.
We develop a set of hard codes to generate $7$-DoF control parameters.
For motion-based skills, we interpolate a trajectory from current position of robot arm to the predicted 3D point from motion planner, and use hard code to move robot arm along this trajectory.
And for primitive skills that interacting with diverse objects (\emph{i.e.}, \textit{pick}, \textit{push}, \textit{pull} and \textit{press}), we individually train primitive-level ACTs~\cite{zhao2023act} as policy-based controllers.
We use $7$-DoF control sequences of corresponding clips to train each policy-based controller.

\vspace{+0.5mm}
\noindent\textbf{Training and Inference Details.}
All parameters in LLaVA, except for vision encoder CLIP~\cite{radford2021clip}, are fine-tuned.
Motion planner, \emph{i.e.}, DETR, in \agent{} is jointly trained with LLaVA.
We train the \agent{} on 8 NVIDIA A100 GPU.
Besides, to adapt DETR to new environment characterized by different sensors during inference, we collect an small dataset from evaluation environment to fine-tuning DETR alone after joint-training while keep VLM frozen.
During inference, we deploy task planner, motion planner and controllers in \agent{} on a NVIDIA A100 GPU, and develop a communication module between \agent{} and robot arms.
The whole inference pipeline operates as depicted in Section~\ref{subsec:pipeline}.

\begin{figure}[t]
    \centering
    \includegraphics[width=0.9\linewidth]{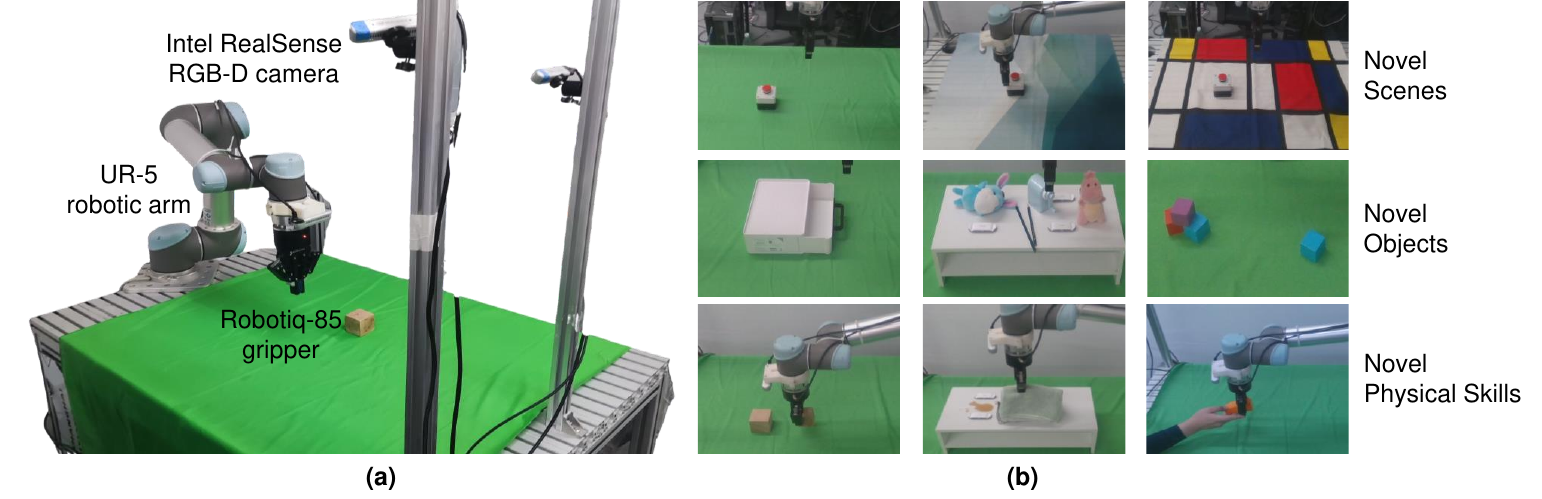}
    \caption{\textbf{(a)} Evaluation platform. \textbf{(b)} Evaluation on generalization of three levels in robotic manipulation tasks.}
    \label{fig:exp_platform}
\end{figure}

\section{Experiments}
\label{sec:exp}

\subsection{Experimental Setup}
\label{subsec:exp_setup}

\noindent\textbf{Evaluation Platform.}
As shown in Figure~\ref{fig:exp_platform} (a), we employ a UR-5 robotic arm with a parallel Robotiq-85 gripper for interaction. 
An Intel RealSense RGB-D camera is positioned in front of robot to capture environmental information. 

\vspace{+0.5mm}
\noindent\textbf{Evaluation setup.}
We select 8 out-of-distribution novel tasks to evaluate the generalizability of \agent{} on three levels:
\begin{enumerate}
    \item \textbf{Novel scenes}: tasks are evaluated in unseen scenes, building with different background and tableclothes, including \textit{Pick Object} and \textit{Press Button}. 
    \vspace{+0.5mm}
    \item \textbf{Novel objects \& compositions}: tasks with unseen objects or unseen compositions (consisting of seen skills and seen objects) are tested, including \textit{Take Down Object} and \textit{Close Drawer}.
    \item \textbf{Novel skills}: tasks with unseen skills are tested, including \textit{Wipe Table}, \textit{Throw Garbage}, \textit{Stack Blocks} and \textit{Receive Object}.
\end{enumerate}
Note that these levels are incremental, suggesting that the test for level 3 also includes the test for levels 1 and 2.
Besides, \textit{\textbf{the 8 tasks in the evaluation and the novel scenes/objects/skills appeared in the evaluation are not present in the training distribution of \agent{}.}}

\begin{table*}[tb]
  \caption{
    \textbf{Evaluation of novel tasks on three levels (10 trials).} 
    ``Plan'' denotes planning accuracy and ``Exec.'' denotes the execution success rate of whole system (including task planning and motion planning if exists). 
    ``w/ FT.'' and ``wo/ FT'' denote ACT with/without fine-tuning on evaluation tasks. 
    Note that our \agent{} have not seen evaluation tasks during training.
  }
  \label{tab:main_res}
  \centering
  \resizebox{0.86\linewidth}{!}{
  \begin{tabular}{@{}lcccccc@{}}
    \toprule
    \multirow{2}{*}{}    & \multicolumn{2}{c}{Novel Scenes} & \multicolumn{2}{c}{Novel Objects \& Compositions} & \multicolumn{2}{c}{Novel Skills} \\
    \cmidrule(lr){2-3}\cmidrule{4-5}\cmidrule{6-7}
                  & Plan (\%)    & Exec. (\%)  & Plan (\%)   & Exec. (\%)  & Plan (\%)   & Exec. (\%) \\
    \midrule
    ACT (wo/ FT.) & -            & 10          & -           & 5           & -           & 5 \\
    ACT (w/ FT.)  & -            & 40          & -           & 25          & -           & 15 \\
    GPT-4V        & \textbf{100} & 35          & \textbf{95} & 17.5        & \textbf{95} & 12.5 \\
    \midrule 
    RA-P (ours)   & \textbf{100} & \textbf{80} & 85          & \textbf{70} & 87.5        & \textbf{67.5} \\
    \bottomrule
  \end{tabular}}
\end{table*}

\vspace{+0.5mm}
\noindent\textbf{Comparison Counterparts.}
We choose ACT~\cite{zhao2023act} as comparison for imitation learning.
We provide two ACT baselines, \emph{i.e.}, the first is pre-trained on the entire \data{} dataset, and another additionally collects datasets of each evaluation tasks and individually trains 8 ACTs based on pre-trained weights following RH20T~\cite{fang2023rh20t}.
We also introduce an agent that uses GPT-4V for both task planning and motion planning, representing agents~\cite{hu2023vila,wake2023gpt4robotics} solely relying on VLMs for \term{} due to the lack of primitive-level spatial knowledge.
We use ICL to guide GPT-4V in selecting primitive skills in \data{} and predicting destination for each motion-based skill.
Due to difficulties in injecting external knowledge, GPT-4V predicts the 2D coordinate as a compromise and constructs the $(x,y,d)$ triplet through directly using the depth information of that coordinate in the image.

\vspace{+0.5mm}
\noindent\textbf{Metric.}
We conduct 10 trials for each task to measure \textit{\textbf{execution success rate}}, assessing whether the whole \agent{} system can accomplish the given tasks. 
To evaluate the performance of task planner, we record planning outputs during execution and manually examine whether task planner correctly generates primitives and objects (\textit{\textbf{planning accuracy}}).
Note that execution success rates contain evaluations of both task planning and motion planning.
And we leave the assessment of motion planning in execution phase for potential discrepancies, \emph{i.e.}, a seemingly feasible destination in planning stage may lead to failure in low-level execution. 

\subsection{Experimental Results}
\label{subsec:exp_result}

The results are shown in Table~\ref{tab:main_res} and Figure~\ref{fig:exp_platform} (b). 

\vspace{+0.5mm}
\noindent\textbf{Comparison to Imitation-based Methods.}
ACT exhibits consistently poor performance, even in the basic tasks like \textit{Pick Object}.
We find the success rate of ACT decreases when the initial position of the robot arm is far from the target objects.
In contrast, by decoupling motion planning from subsequent execution, the primitive-level controller in \agent{} can perform picking operation near the target object, resulting in a significant performance improvement.
Besides, these basic tasks often serve as components of more complex tasks such as \textit{Stack Block}.
The potential of agents that delegate motion planning to low-level controllers will be limited by the poor outcome of low-level controller.

\begin{figure}
    \centering
    \includegraphics[width=1.0\linewidth]{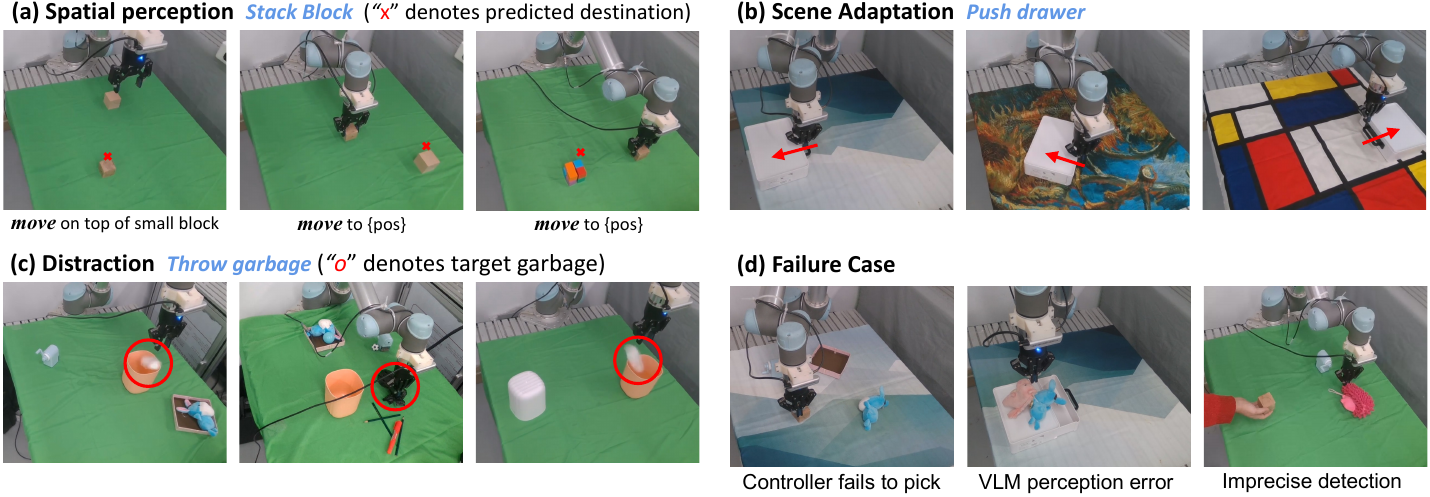}
    \caption{
        \textbf{Visualization of \agent{} executing the tasks.} 
    }
    \label{fig:vis}
    \vspace{-2mm}
\end{figure}

\begin{figure}
    \begin{minipage}{0.55\linewidth}
        \centering
        \includegraphics[width=\linewidth]{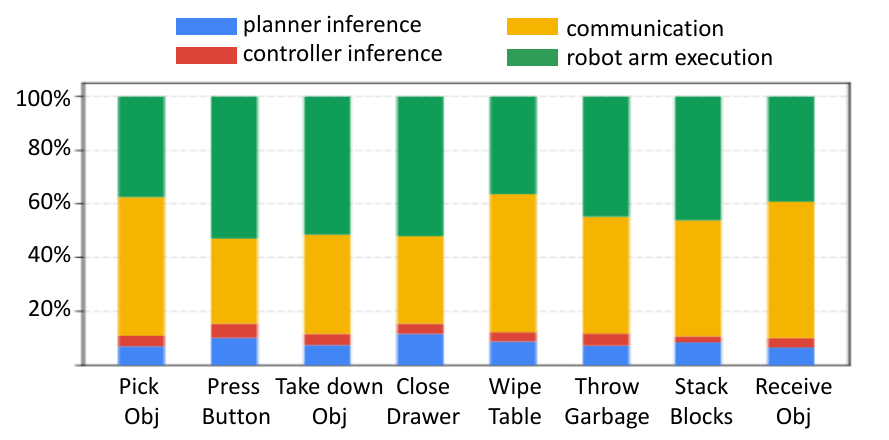}
        \caption{\textbf{Inference time of deployed \agent{}.}}
        \label{fig:inference_time}
    \end{minipage}
    \hspace{0.02\linewidth}
    \begin{minipage}{0.43\linewidth}
        \centering
        \captionof{table}{
            \textbf{Data scaling during fine-tuning.} ``Plan'' and ``Dest.'' denote planning accuracy and mean recall of predicted destination (threshold $\le$ 5cm/10cm), respectively.
        }
        \resizebox{\linewidth}{!}{
            \begin{tabular}{*{4}{c}}
                \toprule
                VLMs                                               & Data   & Plan           & Dest. (5cm/10cm) \\
                \midrule
                \multirow{4}{*}{\shortstack{\agent{}\\(LLaVA-7B\\+ DETR)}} & -      & 0.45     & 0.10/0.31 \\
                                                                   & 20\%   & 0.71     & 0.26/0.60 \\
                                                                   & 50\%   & 0.79     & 0.38/0.72 \\
                                                                   & 100\%  & 0.86     & \textbf{0.48/0.78} \\
                \midrule
                GPT-4V                                             & -      & \textbf{0.94}     & 0.02/0.30 \\
                \bottomrule
            \end{tabular}}
        \label{tab:data_volume}
    \end{minipage}
\end{figure}

\vspace{+0.5mm}
\noindent\textbf{Comparison to Agents relying on VLMs for motion planning.}
With larger-scale VLM, agents built by GPT-4V achieves a higher planning accuracy.
However, obtaining reliable spatial information from GPT-4V through ICL poses great challenges, resulting in a huge disparity in the execution success rate of GPT-4V.
In contrast, our \agent{} achieves a higher execution success rate across all three levels through \term{}, especially for novel skills, which can be attributed to the well-designed primitive skills and corresponding spatial information in \data{}.

\subsection{Qualitative Analysis and Discussion}
\label{subsec:analysis}

\noindent\textbf{Visualization.} 
As shown in Figure~\ref{fig:vis}, we provide an illustration of our \agent{} executing some of tasks in Table~\ref{tab:main_res}.
More video demonstrations of \agent{} are provided on an anonymous webpage (https://sites.google.com/view/rh20t-p/main).

\vspace{+0.5mm}
\noindent\textbf{Robustness on Object Distractions.}
As shown in Figure~\ref{fig:vis} (c), we place an object classified as garbage in a pile of unrelated objects and ask \agent{} to conduct the ``Throw garbage'' task.
\agent{} can distinguish the target object from surroundings based on the observations and successfully execute the task, validating the robustness to object distractions.

\vspace{+0.5mm}
\noindent\textbf{Failure Case.} 
As shown in Figure~\ref{fig:vis} (d), failures primarily include in low-level controllers (such as failing to pick a target object), DETR localization deviation (especially in scenarios with distractions), and perception error of VLM, which leads to subsequent incorrect positioning. 

\vspace{+0.5mm}
\noindent\textbf{Data Scaling.} 
To explore the impact of data scaling during fine-tuning, we construct an simple online benchmark without execution.
The results are shown in Table~\ref{tab:data_volume}.
Both task planning and motion planning are significantly improved compared to baseline.
They are far from saturated, leaving potential room for method design and data accumulation.

\vspace{+0.5mm}
\noindent\textbf{Inference Time.}
Inference time of deployed \agent{} is shown in Figure~\ref{fig:inference_time}.
Using a 7B language model as a decision-making backend for inference is acceptable in terms of time ($\sim$ 8\%).
There still leaves room for larger-scale language models.
We will continue to optimize the efficiency of the entire pipeline through asynchronous communication.

\section{Conclusion}
\label{sec:conclusion}
\vspace{-1mm}

In this work, we introduce \data{}, a dataset designed for primitive-level robotic manipulation that features meticulously defined primitive skills and diverse primitive-level spatial knowledge of multiple forms.
We believe \data{} can facilitate the \tabbre{} applications in robotics, especially in acquiring novel skills.
We also present experimental demonstrations based on proposed plan-execute \tabbre{} paradigm that the agent built on \data{} showcases feasibility and robust generalization in real-world robotic manipulation tasks.

\vspace{+0.5mm}
\noindent\textbf{Limitation.} 
While \data{} serves as a pioneering primitive-level robotic manipulation dataset for real-world \tabbre{} applications, empirical studies indicate the great potential for further data accumulation.
By scaling primitive-level dataset, we anticipate advancements in research on \term{}, significantly expanding generalization capabilities in robotic learning.
Additionally, task-planner (LLaVA-7B) and motion planner (DETR) currently used in \agent{} face constraints due to computing resources.
In the future, we will explore more sophisticated planning system like~\cite{wake2023gpt4robotics,ajay2024hip} and robust motion planning strategies based on direction~\cite{nasiriany2024pivot,li2023manipllm} or trajectories~\cite{gu2023rttrajectory,zhi2023learning} in \tabbre{}s.

\newpage

\section*{\Large Appendix}

\appendix

\section{More Details about \agent{}}
\label{sec:details_appdx}

\begin{figure}[th]
    \centering
    \includegraphics[width=0.9\linewidth]{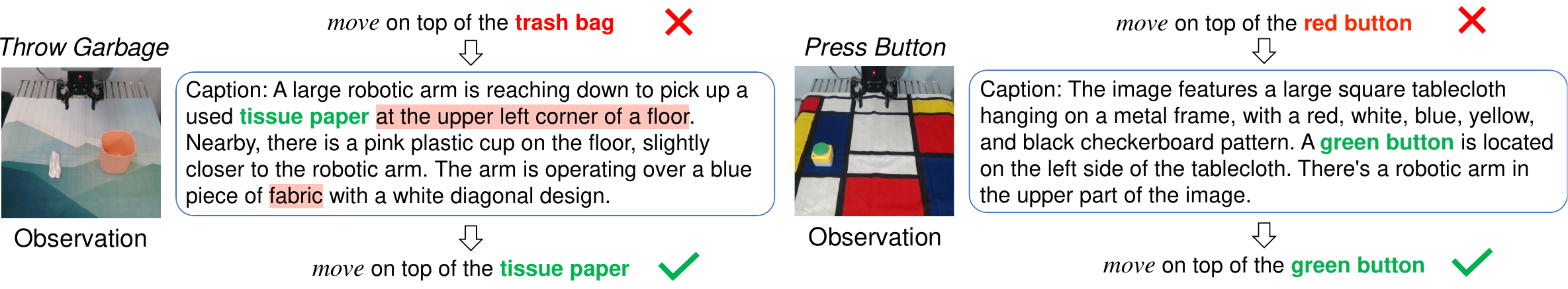}
    \caption{\textbf{{Chain-of-Thought inference.}}}
    \label{fig:cot}
\end{figure}

\begin{table}[th]
\centering
\begin{minipage}{0.99\columnwidth}\vspace{0mm}
\centering
\begin{tcolorbox} 
    \centering
    \small
    \hspace{-1.5mm}
    \begin{tabular}{p{0.99\columnwidth}}
    \begin{minipage}{0.99\columnwidth}\vspace{0mm}

        \VarSty{messages} = \texttt{[\{"role":"system", "content": f"""}
        A chat between a curious user and an artificial intelligence assistant. 
        The assistant is required to guide a robotic arm to perform a complex and comprehensive robotic task. 
        Based on the current observation, completed actions, and current position of the robotic arm, you should give the next action for the robotic arm. \\
        
        The given completed actions and the decision for the next command are composed of the following optional actions: \texttt{[} 
        \begin{itemize}
            \item[$\bullet$] \textit{move} to the \texttt{\{pos\}}
            \item[$\bullet$] \textit{move} \texttt{\{on top of | in front of | ...\}} the \texttt{\{object\}}
            \item[$\bullet$] \textit{push / pull} the \texttt{\{object\}} to the \texttt{\{pos\}}
            \item[$\bullet$] \textit{pick / place} the \texttt{\{object\}}
            \item[$\bullet$] \textit{open / close} the gripper
            \item[$\bullet$] \textit{press} the \texttt{\{object\}} \texttt{\{pos\}}
            \item[$\bullet$] \textit{rotate} \texttt{\{clockwise | counterclockwise\}} \texttt{\{angle\}} \texttt{\{pos\}}
            \item[$\bullet$] \textit{done / reset}
        \end{itemize}
        Here \texttt{\{object\}} and \texttt{\{pos\}} are the templates for the value that needs to be predicted. 
        The position of the robotic arm is represented as [x, y, d] with floating numbers, where x and y denote the coordinates of the robotic arm range from 0 to 1, and d denotes the depth of the robot arm in the observation.
        \texttt{"""\}]}

    \end{minipage}
    \end{tabular}
\end{tcolorbox}
\caption{The system prompt for robotic tasks in \agent{}.}
    \label{tab:sys_prompt_rap_robotic}
\end{minipage}
\end{table}

\noindent\textbf{Chain-of-Thought Inference.}
We find some perceptual errors in classifying object categories and attributes during decision-making, due to the limited distribution of objects in RH20T~\citep{fang2023rh20t} and VLM scale (7B) we used in \agent{}.
Consequently, we propose to use the VLM to first describe the scene and then make decisions based on the generated descriptions.
We refer it as to Chain-of-Thought inference (CoT inference).
No extra adjustments are required during the training phase; instead, we can directly add the descriptions generated by VLM to the prompts for inference.
Note that we only caption the scene before the initial decision, and all subsequent decisions within the same task rely on the same description, thereby increasing a minor overhead to the inference time of planning.
As illustrated in the Figure~\ref{fig:cot}, CoT inference can effectively reduce some perceptual errors thanks to scene descriptions.
Given the scale of model we used in \agent{} (7B), there is still room for improvement in captioning to assist VLM with subsequent decision-making, especially in terms of hallucinations included in the descriptions (texts marked with a red background in Figure~\ref{fig:cot}) and scenarios with multiple object interferences.
Besides, GPT-4V~\citep{openai2023gpt4v} can be used as an option to describe the scene for CoT inference.


\noindent\textbf{Prompts and Instructions used in \agent{}.}
System prompt in \agent{} is listed in Table~\ref{tab:sys_prompt_rap_robotic}.
Instructions for robotic tasks in \agent{} is listed in Table~\ref{tab:inst_rap_robotic}.
Here, \texttt{\{task\_desc\}}, \texttt{\{historical\_decisions\}} and \texttt{\{robot\_arm\_pos\}} denote language specifications like ``Pick Blocks'', historical decisions made by task planners before current state and the position of the robot arm, respectively.
During training, we randomly select one of the instructions in the conversations.

\begin{table}[th]
\centering
\begin{minipage}{0.99\columnwidth}\vspace{0mm}
\centering
\begin{tcolorbox} 
    \centering
    \small
    \hspace{-6mm}
    \vspace{-4mm}
\begin{itemize}[leftmargin=4.5mm]
    \setlength{\itemsep}{2pt}
    \item[$\bullet$] To accomplish the task \texttt{\{task\_desc\}}, the following actions have been sequentially completed: \texttt{\{historical\_decisions\}}. Based on the observation and scene depiction (\texttt{\{scene\_desc\}}), the current position of the robotic arm is \texttt{\{robot\_arm\_pos\}}. What should be the next action of the robotic arm?
    \item[$\bullet$] Given the embodied task \texttt{\{task\_desc\}}, the subsequent steps have been undertaken: \texttt{\{historical\_decisions\}}. Referring to the observation and scene description (\texttt{\{scene\_desc\}}), the present location of the robotic arm is indicated by \texttt{\{robot\_arm\_pos\}}. What would be the appropriate next action for the robotic arm?
    \item[$\bullet$] With the assigned task \texttt{\{task\_desc\}}, the following actions have been accomplished: \texttt{\{historical\_decisions\}}. As shown in the observation and the description (\texttt{\{scene\_desc\}}), the current position of the robotic arm is \texttt{\{robot\_arm\_pos\}}. Considering this, what are the next advisable action for the robotic arm?
    \item[$\bullet$] In light of the assigned task \texttt{\{task\_desc\}} and corresponding description (\texttt{\{scene\_desc\}}), the ensuing steps were carried out: \texttt{\{historical\_decisions\}}. Observing the image given, the robotic arm\'s current position is marked as \texttt{\{robot\_arm\_pos\}}. In light of this, what is the recommended next step for the robotic arm?
    \item[$\bullet$] For the purpose of achieving the goal of \texttt{\{task\_desc\}}, we have progressively executed these steps: \texttt{\{historical\_decisions\}}. As indicated by the observation and scene depiction (\texttt{\{scene\_desc\}}), the robotic arm\'s current location is \texttt{\{robot\_arm\_pos\}}. What would be the advisable subsequent action for the robotic arm?
    \item[$\bullet$] To achieve the goal of \texttt{\{task\_desc\}}, we have progressively executed these steps: \texttt{\{historical\_decisions\}}. As indicated by the observation and scene depiction (\texttt{\{scene\_desc\}}), the robotic arm\'s current location is \texttt{\{robot\_arm\_pos\}}. What would be the advisable subsequent action for the robotic arm?
    \item[$\bullet$] To reach the objective of \texttt{\{task\_desc\}}, we have methodically performed the following actions: \texttt{\{historical\_decisions\}}. The observation shows that the robotic arm is currently positioned at \texttt{\{robot\_arm\_pos\}}. Given the scene description (\texttt{\{scene\_desc\}}), what should be the next action for the robotic arm?
    \item[$\bullet$] Aimed at accomplishing the goal of \texttt{\{task\_desc\}}, we have systematically undertaken these steps: \texttt{\{historical\_decisions\}}. As depicted in the supplied image, the location of the robotic arm is currently \texttt{\{robot\_arm\_pos\}}. Considering this, what would be the prudent next step for the robotic arm?
    \item[$\bullet$] With the given task \texttt{\{task\_desc\}}, a sequence of actions has been diligently executed: \texttt{\{historical\_decisions\}}. The current position of the robotic arm, as shown in the image, is \texttt{\{robot\_arm\_pos\}}. Based on the scene depiction (\texttt{\{scene\_desc\}}), what would be the logical next action for the robotic arm?
    \item[$\bullet$] Considering the ongoing \texttt{\{task\_desc\}}, our approach has involved a series of progressive steps: \texttt{\{historical\_decisions\}}. The image provided points out the current position of the robotic arm at \texttt{\{robot\_arm\_pos\}}. The scene description is \texttt{\{scene\_desc\}}. What is the next recommended action for the robotic arm in this scenario?
\end{itemize}
\end{tcolorbox}
\caption{The list of instructions used for robotic tasks in \agent{}.}
    \label{tab:inst_rap_robotic}
\end{minipage}
\end{table}

\begin{figure}[tb]
    \centering
    \includegraphics[width=0.95\linewidth]{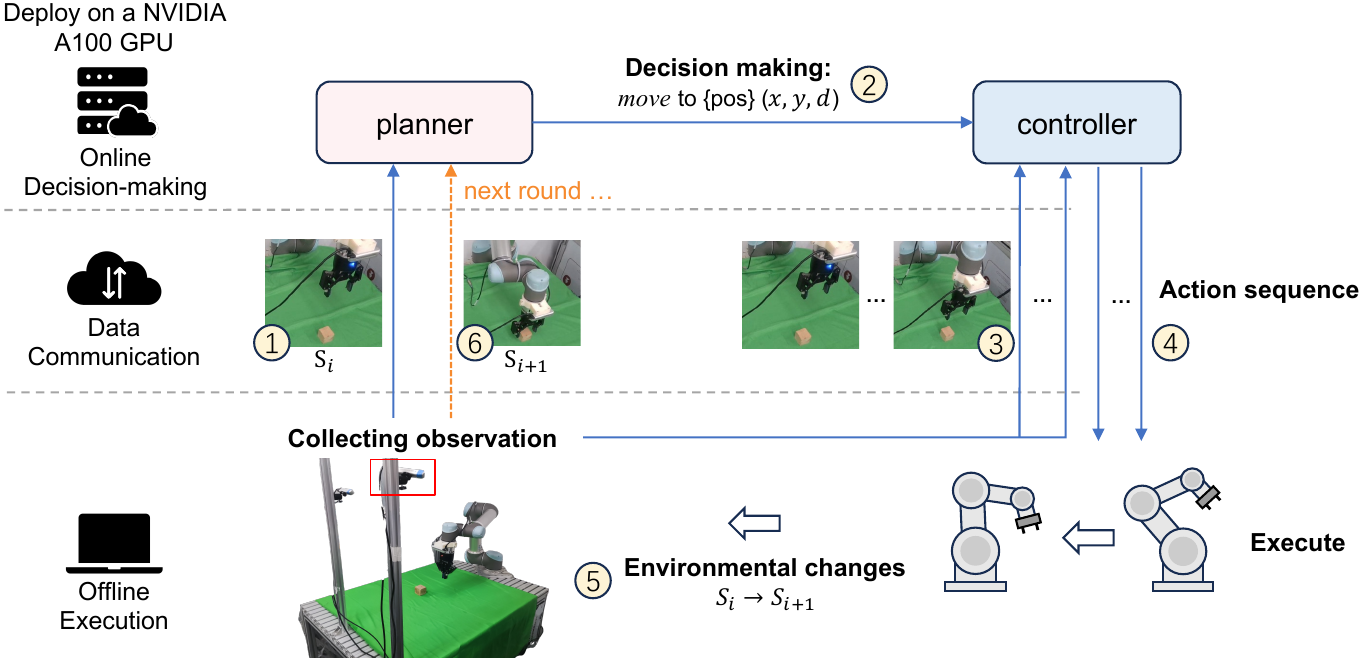}
    \caption{
        \textbf{Deployment pipeline of \agent{}.} 
    }
    \label{fig:deploy_pipe}
\end{figure}

\noindent
\textbf{Execution Deployment.}
As shown in Figure~\ref{fig:deploy_pipe}, we deploy LLaVA, Deformable DETR~\citep{zhu2020deformabledetr} and low-level ACT~\citep{zhao2023act} controllers on a NVIDIA A100 GPU, and develop a communication module between \agent{} and robot arm.
We conduct the following procedures to perform the plan-execute paradigm during the evaluation stage:
\begin{enumerate}
    \item \textbf{Collecting observation:} 
    the sensor in the evaluation platform captures the observation information in the environment, and then transmits the RGB images along with the current position of the robot arm to the agent, which is deployed on an A100 GPU, through the communication module.
    \item \textbf{Decision making:} 
    the task planner will take images and position them as input and make decisions using predefined primitive skills. 
    If any motion-based skill is chosen, the motion planner will be invoked to predict the precise coordinate of the destination $(x,y,d)$ where the robot arm should move to. 
    The coordinate will be transformed into a 3D coordinate with camera calibration.
    \item \textbf{Mapping to action sequence and then executing:} 
    based on the type of primitive skills, a specific low-level controller will be called to map the decision and coordinate to the action sequence.
    For policy-based controllers, we predict action sequences based on current observations for the next 5 steps.
    After receiving and executing the 5-step action sequence, the robot arm collects information again and transmits it to controllers, repeating until the controllers give a terminate signal.
    For hard-code controllers, we interpolate a straight line between the starting position of the robot arm and the predicted destination to obtain a movement trajectory, then use hard code to move the robot arm along the trajectory as the action sequence.
    The robot arm then executes the action sequence, omitting the multiple data transfers and communications like those in policy-based controllers.
    \item \textbf{Iterative plan-execute process:} 
    Once the controllers complete the decision made by the planners, we will return to step 1 to start a new round of the plan-execute process until the planner ultimately gives a \textit{done} decision.
\end{enumerate}

We also provide a analysis on inference time in Section~\ref{subsec:analysis}.

\begin{table}[t]
    \centering
    \begin{minipage}{0.99\columnwidth}\vspace{0mm}
    \centering
    \begin{tcolorbox} 
        \centering
        \small
        \hspace{-1.5mm}
        \begin{tabular}{p{0.99\columnwidth}}
        \begin{minipage}{0.99\columnwidth}\vspace{0mm}
    
            \VarSty{messages} = \texttt{[\{"role":"system", "content": f"""}
            You are placed inside an embodiment environment with a robot arm and a gripper with it. 
            Given a goal (task description) in language and multi-frame RGB images depicting the scene and objects, you are required to decompose the goal into sub-tasks step by step to ensure it can be accomplished eventually. \\
            
            At each round of conversation, I will give you:
            \begin{itemize}
                \item[$\bullet$] multi-frame RGB images, time from past to present
                \item[$\bullet$] task description in language
                \item[$\bullet$] historical decisions already made
                \item[$\bullet$] The relative coordinates (x, y, d) in the current image where the gripper is located. 
                The x-axis is from left to right, and the y-axis is from top to bottom, with a range of values between 0 and 1. 
                For example, the top left corner of the image is (0,0), and the top right corner is (1,0).
            \end{itemize}
    
            Your reply should be one of the primitive skills, as follows:
            \begin{itemize}
                \item[$\bullet$] \textit{move} to the \texttt{\{pos\}}
                \item[$\bullet$] \textit{move} \texttt{\{on top of | in front of | ...\}} the \texttt{\{object\}}
                \item[$\bullet$] \textit{push / pull} the \texttt{\{object\}} to the \texttt{\{pos\}}
                \item[$\bullet$] \textit{pick / place} the \texttt{\{object\}}
                \item[$\bullet$] \textit{open / close} the gripper
                \item[$\bullet$] \textit{press} the \texttt{\{object\}} \texttt{\{pos\}}
                \item[$\bullet$] \textit{rotate} \texttt{\{clockwise | counterclockwise\}} \texttt{\{angle\}} \texttt{\{pos\}}
                \item[$\bullet$] \textit{done / reset}
            \end{itemize}
            Remember to replace the \texttt{\{pos\}} in primitive skills with relative coordinates (x, y) as the position in the image. \\

            I will then give an example to help you follow the context:
            ... \\
            \texttt{"""\}]}
    
        \end{minipage}
        \end{tabular}
    \end{tcolorbox}
    \caption{The system prompt for Agents with GPT-4V in Execution Phase.}
        \label{tab:sys_prompt_gpt4v}
    \end{minipage}
\end{table}
    
\section{GPT-4V Execution Setup}
\label{sec:gpt4v_exec}

\noindent\textbf{System Prompts for agents with GPT-4V during Execution Phase.}
We evaluate agents with GPT-4V through in-context learning. 
Detailed system prompt is shown in Table~\ref{tab:sys_prompt_gpt4v}.

\section{More Results}

\begin{figure}
    \centering
    \includegraphics[width=0.83\linewidth]{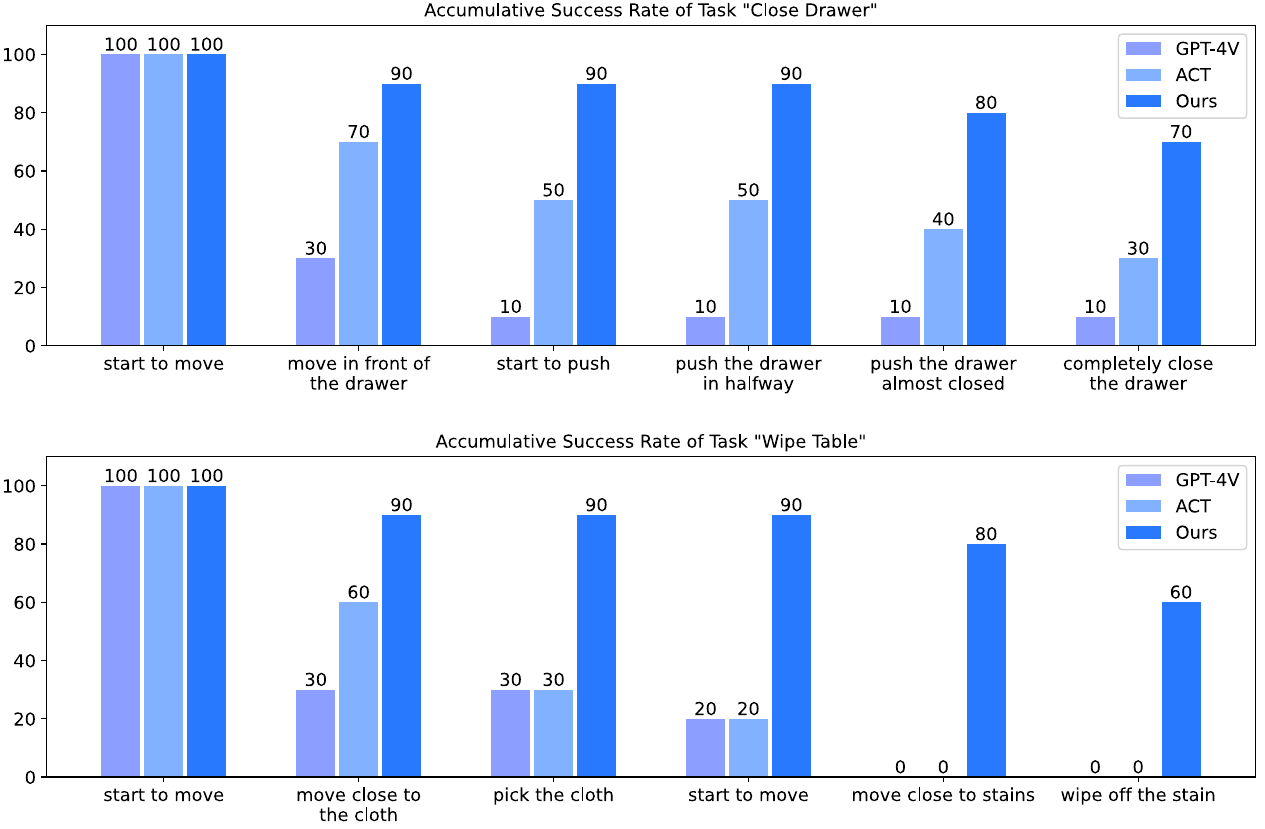}
    \caption{
        \textbf{Cumulative success rates for different stages of several tasks.}
    }
    \label{fig:cum_success}
\end{figure}

\noindent\textbf{Cumulative Success Rates.}
We show the cumulative success rates for different steps of several tasks in Figure~\ref{fig:cum_success}.
It is observed that the majority of task failures in agents with GPT-4V are attributed to insufficient localization abilities.
The positions predicted by GPT-4V, which are far from the target object, result in the failure of the low-level controller's execution.
In contrast, our \agent{} trained on \data{} can provide more reasonable spatial priors, resulting in a higher execution success rate.


\section{All Tasks in \data{}}

We provide the list of tasks in \data{} in Table~\ref{tab:task_list_1}.

\section{Potential Social Impact}
\label{sec:social_impact}

The proposed \data{} dataset and \agent{} model demonstrate effectiveness and generalization in robotic tasks, especially in novel physical skills, which can benefit the future development of \tabbre{}s.
The violent elements (\emph{e.g.}, using a knife) in the dataset and related knowledge learned by the robot may have some potential negative social impacts.
However, considering that our data source, RH20T, has already been publicly released, these impacts are controllable.




\begin{figure}
    \centering
    \includegraphics[width=0.78\linewidth]{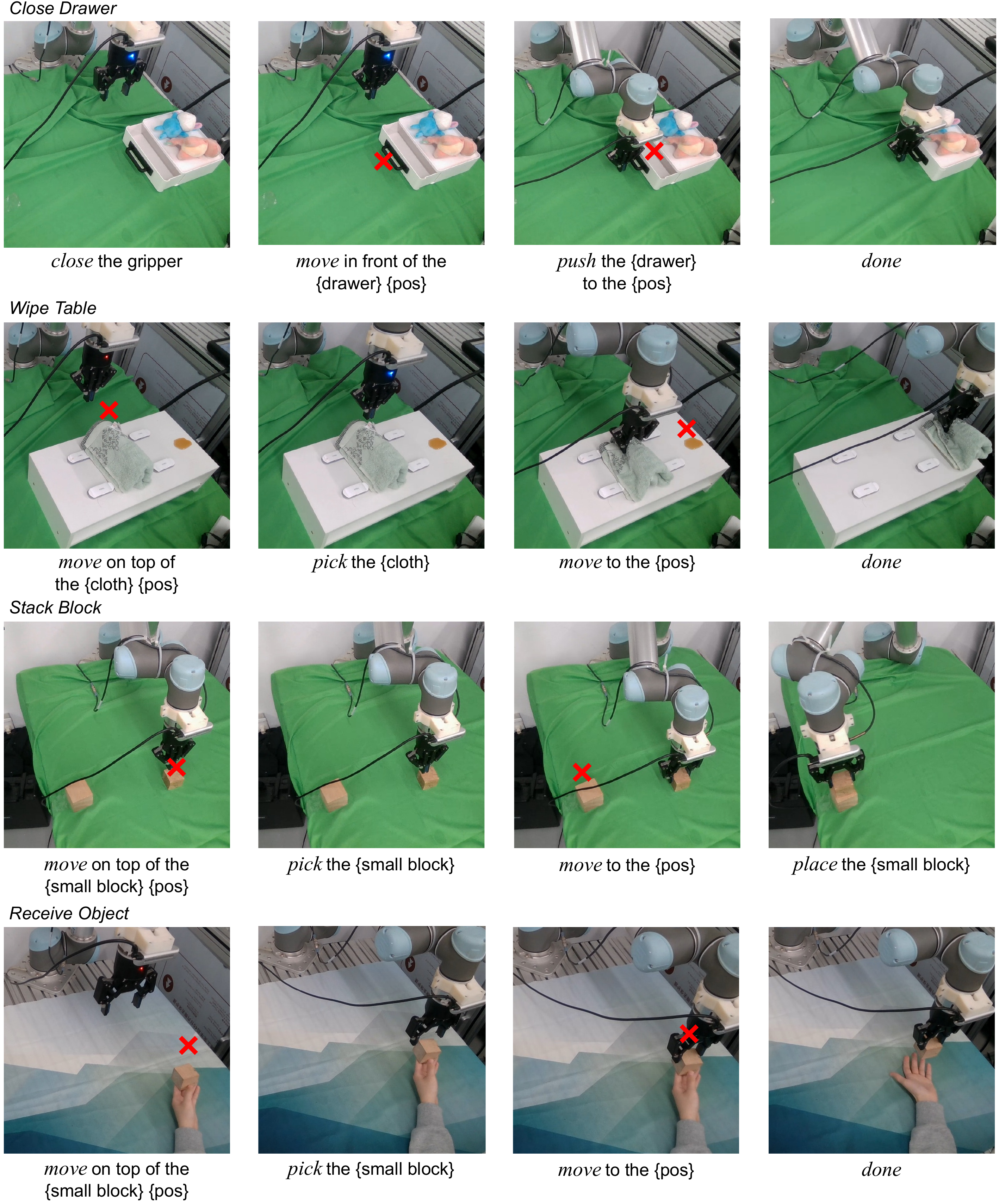}
    \caption{
        \textbf{Complete execution process of \agent{} during evaluation.}
    }
    \label{fig:execution_vis}
\end{figure}

\begin{table}
\centering
\begin{minipage}{0.99\columnwidth}\vspace{0mm}
\centering
\begin{tcolorbox} 
    \centering
    \small
    \hspace{-6mm}
    \vspace{-4mm}
\begin{itemize}[leftmargin=4.5mm]
    \setlength{\itemsep}{2pt}
    \item[$\bullet$] Turn off the desk lamp
    \item[$\bullet$] Take the dish off the dish rack
    \item[$\bullet$] Turn the knob to increase the volume of the speaker
    \item[$\bullet$] Placing a piece on the chessboard to complete the setup.
    \item[$\bullet$] Clean the table with a cloth
    \item[$\bullet$] Throw the garbage
    \item[$\bullet$] Take the pencil out from the pencil sharpener
    \item[$\bullet$] Pull out a napkin
    \item[$\bullet$] Plug in the power cord of the desk lamp, turn on the socket, and light up the desk lamp
    \item[$\bullet$] Press three buttons from left to right in sequence
    \item[$\bullet$] Grab the block and place it at the designated location
    \item[$\bullet$] Turn the hands of a clock
    \item[$\bullet$] Remove the object from the scale
    \item[$\bullet$] Play the first move as black on the 3-4 points in the upper right corner of the Go board
    \item[$\bullet$] Grasp the handle and open the drawer
    \item[$\bullet$] Remove the bubble ring from the assembled bubble ring and ball
    \item[$\bullet$] Pick up one small block
    \item[$\bullet$] Take the photo frame down from the bracket
    \item[$\bullet$] Put the object on the shelf
    \item[$\bullet$] Pick up and place an object with obstacles
    \item[$\bullet$] Put the knife on the cutting board
    \item[$\bullet$] Press the button
    \item[$\bullet$] Approach and touch the side of the small block
    \item[$\bullet$] Hold a block with the gripper and sweep it from left to right
    \item[$\bullet$] Close the drawer
    \item[$\bullet$] Take everything out of the gift box
    \item[$\bullet$] Press a button from top to bottom with obstacles
    \item[$\bullet$] Open the microwave door
    \item[$\bullet$] Turn on the water tap
    \item[$\bullet$] Place the handset of the telephone on the corresponding phone cradle
    \item[$\bullet$] Play the drum
    \item[$\bullet$] Turn on the desk lamp by pressing the button
    \item[$\bullet$] Turn on the power strip by pressing the button
    \item[$\bullet$] Put the cup on the cup rack
    \item[$\bullet$] Open a sliding window
    \item[$\bullet$] Move an object from one box to another
    \item[$\bullet$] Cover the pot with the lid
    \item[$\bullet$] Push an object with obstacles
    \item[$\bullet$] Use the gripper to push the small block from left to right
    \item[$\bullet$] Pick up the cup
    \item[$\bullet$] Wipe the tabletop with a sponge
    \item[$\bullet$] Take the object down from the shelf
    \item[$\bullet$] Push down the lever
    \item[$\bullet$] Turn off the water tap
\end{itemize}
\end{tcolorbox}
\caption{The list of tasks in \data{}.}
    \label{tab:task_list_1}
\end{minipage}
\end{table}

\end{document}